\documentclass[conference]{IEEEtran}
\IEEEoverridecommandlockouts

\usepackage[normalem]{ulem}

\usepackage{soul}
\usepackage[compatibility=false]{caption}
\usepackage{amsmath,amsfonts}
\usepackage{algorithmicx}  
\usepackage{algpseudocode}
\usepackage{array}
\usepackage[caption=false,font=normalsize,labelfont=sf,textfont=sf]{subfig}
\usepackage{stfloats}
\usepackage{url}
\usepackage{verbatim}
\usepackage{graphicx}
\usepackage{hyperref}

\hypersetup{
    colorlinks=false, 
    urlcolor=magenta,   
}
\usepackage{pbox}
\usepackage{algorithm}
\usepackage{multirow}
\usepackage{bbding}
\usepackage{array}
\usepackage{makecell}
\usepackage{xspace}
\usepackage{color}
\usepackage{xcolor}
\usepackage{amssymb}
\usepackage{booktabs}
\usepackage{pifont}
\usepackage{bm}

\def\eg{\textit{e.g.}}
\def\ie{\textit{i.e.}}

\def\ours{\emph{ReMA}\xspace}
\def\MA{\emph{RAM}\xspace}
\def\MB{\emph{DSM}\xspace}

\usepackage{cite}
\usepackage{booktabs,multirow}
\usepackage{pifont}        
\usepackage{caption}       
\usepackage{subcaption}    

\def\BibTeX{{\rm B\kern-.05em{\sc i\kern-.025em b}\kern-.08em
    T\kern-.1667em\lower.7ex\hbox{E}\kern-.125emX}}
\begin{document}

\title{\emph{ReMA}: A Training-Free Plug-and-Play Mixing Augmentation for Video Behavior Recognition}
\author{
\IEEEauthorblockN{Feng-Qi Cui\textsuperscript{1,3}, Jinyang Huang\textsuperscript{2,†}, Sirui Zhao\textsuperscript{1}, Jinglong Guo\textsuperscript{2}, Qifan Cai\textsuperscript{2,3}, Xin Yan\textsuperscript{4}, Zhi Liu\textsuperscript{5} }

\IEEEauthorblockA{\textit{\textsuperscript{1} University of Science and Technology of China}, Hefei, China 
 \quad
\textit{\textsuperscript{2} Hefei University of Technology}, Hefei, China\\
\textit{\textsuperscript{3} Hefei Xiaosheng Intelligent Technology Co., Ltd.}, Hefei, China \quad
\textit{\textsuperscript{4} Cylingo Group},
Beijing, China \\
\textit{\textsuperscript{5} The University of Electro-Communications,}
Tokyo, Japan \\
\textsuperscript{†}Corresponding Author: hjy@hfut.edu.cn }

}

\maketitle

\begin{abstract}
Video behavior recognition demands stable and discriminative representations under complex spatiotemporal variations.
However, prevailing data augmentation strategies for videos remain largely perturbation-driven, often introducing uncontrolled variations that amplify non-discriminative factors, which finally weaken intra-class distributional structure and representation drift with inconsistent gains across temporal scales. To address these problems, we propose Representation-aware Mixing Augmentation (\ours), a plug-and-play augmentation strategy that formulates mixing as a controlled replacement process to expand representations while preserving class-conditional stability.
\ours integrates two complementary mechanisms.
Firstly, the Representation Alignment Mechanism (\MA) performs structured intra-class mixing under distributional alignment constraints, suppressing irrelevant intra-class drift while enhancing statistical reliability.
Then, the Dynamic Selection Mechanism (\MB) generates motion-aware spatiotemporal masks to localize perturbations, guiding them away from discrimination-sensitive regions and promoting temporal coherence.
By jointly controlling how and where mixing is applied, \ours improves representation robustness without additional supervision or trainable parameters.
Extensive experiments on diverse video behavior benchmarks demonstrate that \ours consistently enhances generalization and robustness across different spatiotemporal granularities.
\end{abstract}


\begin{figure*}[t]
    \centering
    \begin{minipage}[c]{0.63\textwidth} 
        \centering
        \includegraphics[width=1.0\textwidth]{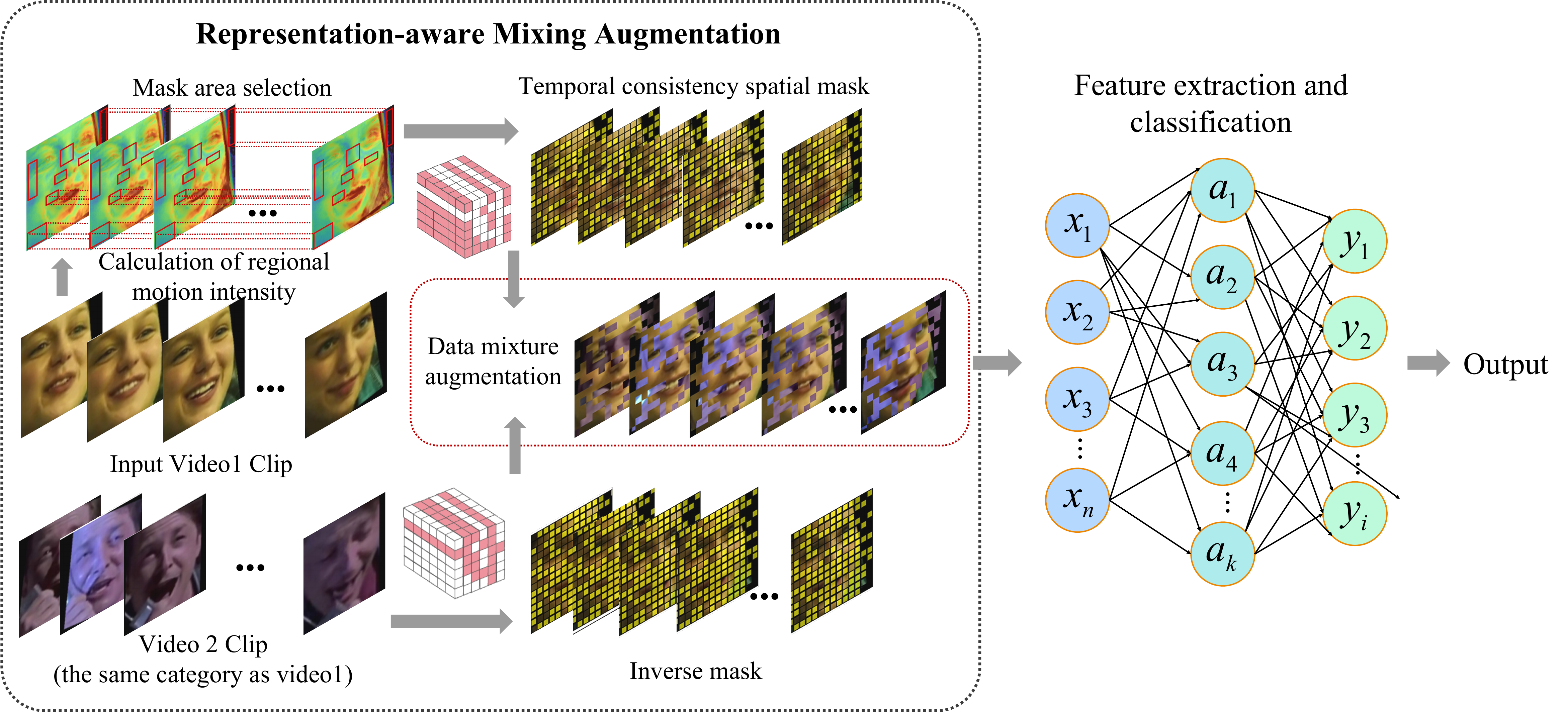}
        \caption{Overview of the proposed \ours. \ours~performs motion-guided, temporally consistent spatial masking and intra-class mixing.}
        \label{Fig_framework}
    \end{minipage}%
    \hfill
    \begin{minipage}[c]{0.35\textwidth} 
        \centering
        \small 
        
        \hrule height 0.8pt depth 0pt \relax
        \vspace{2pt}
        \captionof{algorithm}{The flow of the \ours}
        \label{alg:rema}
        \hrule height 0.4pt depth 0pt \relax
        
    \begin{algorithmic}[1]
        \Require Video dataset $\mathcal{D}$, number of frames $T$, coverage ratio $r$, block size $b_0$
        \Ensure Augmented video $\tilde{x}$
            
        \State Sample $(x_i,y_i)\sim\mathcal{D}$ and uniformly sample $T$ frames $x=\{x_t\}_{t=1}^{T}$
        \State Sample intra-class video $x_j \sim p(x\mid y=y_i)$ and uniformly sample $T$ frames $x'$
        \State Compute motion map $\mathcal{A} \leftarrow \frac{1}{3(T-1)}\sum_{t=1}^{T-1}\sum_{c=1}^{3}\big|x_{t+1}^{(c)}-x_t^{(c)}\big|$
        \State Pool $\mathcal{A}$ into patch-level motion map $\mathcal{P}$ with block size $b_0$
        \State Sample $r$-ratio patches according to $p_{ij}\propto (1-\mathcal{P}_{ij})$
        \State Construct tube-consistent mask $\mathcal{M}$ shared across all frames
        \State Mix videos by $\tilde{x}=(1-\mathcal{M})\odot x + \mathcal{M}\odot x'$
        \State \Return $\tilde{x}$
    \end{algorithmic}
        \hrule height 0.8pt depth 0pt \relax
    \end{minipage}
\end{figure*}

\section{Introduction}
Video behavior recognition is a fundamental problem in computer vision, with broad applications in human-computer interaction, psychological analysis~\cite{cui2025disentanglingemotionalbasestransient}, and intelligent healthcare.
As research progresses, behavior understanding has extended from coarse-grained body actions to finer semantic levels such as facial expressions~\cite{10.1145/3746027.3755036} and subtle micro-actions~\cite{guo2024benchmarking}.
Across these tasks, models are expected to learn discriminative representations that generalize across varying temporal scales, spatial resolutions, and motion patterns~\cite{fan2021multiscale}.
However, a key challenge in current video behavior recognition pipelines lies not in the inherent instability of video representations, but in the lack of explicit control over how training perturbations shape the learned representation space.
Unfortunately, under prevalent data augmentation paradigms, non-discriminative variations are often indiscriminately introduced and progressively amplified, which inevitably undermines the formation of stable and task-relevant decision boundaries~\cite{zou2023video_aug_invariance,11175550}. Specifically, when augmentation effects are not properly constrained, uncontrolled region replacement or interpolation may distort temporal dynamics~\cite{qian2020spatiotemporal_contrastive}, excessive execution-style variation can induce dispersion or drift of class-conditional representations~\cite{wang2020temporal_shift}, and random or fixed-scale perturbations tend to amplify high-frequency redundancy and local noise in the feature space~\cite{10219705}.
Together, these factors cause non-discriminative variations to dominate representation learning, yielding fragile decision boundaries under diverse scenes and motion scales~\cite{fan2023motion_guided_masking}.

Data augmentation is widely adopted to improve generalization~\cite{10310260}.
Beyond standard image transforms~\cite{shorten2019survey}, mixed augmentations, \eg, Mixup~\cite{zhang2017mixup} and CutMix~\cite{yun2019cutmix} expanded training distributions via interpolation or regional replacement~\cite{10219853}.
However, directly extending these strategies to videos often disrupts temporal coherence and motion semantics due to the absence of structure-aware constraints~\cite{cao2024survey_mixda}.
Although video-specific mixed strategies have shown effectiveness in action recognition~\cite{gowda2022learn2augment}, they basically emphasize diversity expansion while overlooking intra-class statistical consistency, producing augmented samples that deviate from the original distribution~\cite{zou2023video_aug_invariance}.
Moreover, without explicit awareness of the corresponding motion structure, perturbations may be applied to critical dynamic regions, which is particularly detrimental to fine-temporal behaviors~\cite{10219625}.

Consequently, the limitation of existing approaches is less about sample scarcity than about insufficient control over the distributional and temporal side effects of perturbations.
Many methods remain diversity-driven, relying on random or fixed perturbations with largely unregulated impact on representation learning~\cite{9936613}.
Fortunately, recent evidence from self-supervised video learning further motivates a control-oriented perspective, \ie, videos exhibit strong spatiotemporal redundancy and correlation, allowing effective representations to be learned even under high masking ratios~\cite{tong2022videomae}.
This observation clearly suggests that a large \textbf{replaceable} or \textbf{perturbable} space exists in videos, and the key lies in exploiting it in a controlled manner.
Thus, \uline{we argue that mixed augmentation should shift from indiscriminate perturbation to controlled replacement, acting primarily on replaceable redundancy while avoiding the degradation of discriminative dynamics and intra-class statistics}.
From a mechanistic standpoint, current methods leave substantial headroom because they rarely constrain whether augmented samples remain within effective class-conditional regions or enforce structure-guided, temporally coherent perturbations aligned with motion-sensitive regions~\cite{10219625}.
To better unlock the potential of mixed augmentation, we propose \textbf{\textit{Representation-aware Mixing Augmentation} (\ours)}, \uline{a controlled mixed video data augmentation method that can be seamlessly integrated into general video backbones.
Rather than treating mixing as an indiscriminate perturbation, \ours explicitly formulates it as an information replacement process and introduces hierarchical control over its distributional and structural effects}.
Specifically, the \textbf{\textit{Representation Alignment Mechanism} (\MA)} performs structured intra-class mixing under distributional alignment constraints, effectively ensuring that augmented samples remain statistically consistent with the original class distribution while suppressing irrelevant intra-class drift.
Complementarily, the \textbf{\textit{Dynamic Selection Mechanism} (\MB)} adaptively generates spatiotemporal masks based on inter-frame motion intensity, guiding perturbations toward relatively less discrimination-sensitive regions and maintaining temporal continuity via time-consistent masking~\cite{tong2022videomae}.
By jointly regulating how and where mixing is applied, \ours improves the controllability and stability of mixed augmentation at the data level, without introducing additional supervision or trainable parameters.

Totally, the main contributions are summarized as follows:
\begin{itemize}
\item We revisit mixed video augmentation from a controlled replacement perspective, motivated by the spatiotemporal redundancy of videos and the uneven distribution of discriminative behavioral cues. Based on this view, we propose \ours, \textbf{a plug-and-play controlled mixing strategy that explicitly regulates the representation-level impact of augmentation without additional supervision or trainable parameters}.
\item We propose the \MA, which performs structured intra-class mixing under distributional alignment constraints to suppress non-discriminative intra-class variations and improve the statistical reliability of augmented samples, enabling more consistent representation learning gains.
\item We propose the \MB, which adaptively generates spatiotemporal masks from inter-frame motion intensity while enforcing temporal consistency, guiding perturbations away from motion-sensitive regions to preserve critical dynamics while expanding effective feature diversity.
\item Extensive experiments are conducted across multiple benchmarks with different behavioral granularities, demonstrating that \ours consistently improves performance and revealing the complementary roles of distributional alignment and structure-aware control in raising the effectiveness ceiling of mixed augmentation.
\end{itemize}

\section{ A New Data Augmentation Method}
We propose \ours, a representation-aware mixing augmentation framework for video behavior recognition, which aims to expand intra-class diversity under controlled spatiotemporal perturbations. The key idea is to reformulate mixing-based augmentation as a class-conditional and structure-consistent replacement process, so that augmented samples can stably contribute to discriminative representation learning. To this end, \ours consists of two complementary mechanisms: the Representation Alignment Mechanism (\MA), which regulates the statistical behavior of intra-class mixing, and the Dynamic Selection Mechanism (\MB), which adapts the spatial locations and granularity of replacement according to video motion characteristics. Both mechanisms operate entirely at the data level and introduce no additional supervision or trainable parameters. The overall augmentation pipeline of \ours is illustrated in Fig. \ref{Fig_framework} and Alg. \ref{alg:rema}.

\subsection{Representation Alignment Mechanism}

\MA aims to improve the statistical stability of mixing-based augmentation, so that augmented samples can contribute more reliably to video representation learning.
Although intra-class mixing is semantically valid, its effectiveness critically depends on whether both the perturbation source and the replacement budget are explicitly constrained.
Without such constraints, mixing operations with varying scales may introduce inconsistent perturbation strength across samples, leading to unstable representation learning.

In \ours, mixing augmentation is formulated as a class-conditional and budget-controlled replacement process.
Given two video samples $x_i$ and $x_j$ from the same class, the augmented sample is constructed via a spatiotemporal mask $\mathcal{M}$, which can be expressed as:
\begin{equation}
\tilde{x} = (1 - \mathcal{M}) \odot x_i + \mathcal{M} \odot x_j,
\quad x_j \sim p(x \mid y = y_i).
\end{equation}
The class-conditional sampling constraint ensures that the replacement content originates from the same category, restricting the perturbation to remain within the semantic scope of intra-class variation.
Let $\Phi(\cdot)$ denote a fixed video feature mapping.
Under this constraint, the augmented samples are statistically distributed around the typical representation region of the corresponding class, such that mixing does not induce systematic class-level distributional drift:
\begin{equation}
\mathbb{E}[\Phi(\tilde{x}) \mid y_i] \approx \mathbb{E}[\Phi(x) \mid y_i].
\end{equation}
As a result, mixing augmentation primarily expands the intra-class feature support instead of shifting the class center.

Beyond constraining the perturbation source, \MA further regulates the replacement budget through the spatiotemporal mask.
Specifically, we define the average coverage ratio as
\begin{equation}
r = \frac{1}{THW} \sum_{t,h,w} \mathcal{M}(t,h,w),
\end{equation}
which specifies the proportion of content replaced in the video.
By fixing $r$, \MA ensures that each augmented sample undergoes a comparable level of perturbation, preventing excessive or insufficient replacement caused by unbalanced mask coverage.
This budget control does not enforce similarity in content differences between samples, but instead standardizes the overall strength of intra-class mixing across the dataset.

Through the above design, \uline{\MA introduces no additional supervision or trainable parameters, yet effectively provides an explicit statistical constraint for mixing-based augmentation}.
To establish a stable foundation for the subsequent motion-guided dynamic selection mechanism, by transforming sample-level mixing into a budget-controlled intra-class expansion in representation space, \MA enables the augmented samples to participate more consistently in learning discriminative decision boundaries. 

\subsection{Dynamic Selection Mechanism}
\MB regulates where mixing augmentation is applied, so that the budget-controlled replacement defined by \MA remains consistent with the spatiotemporal structure of video data.
While \MA constrains the perturbation source and overall replacement budget $r$, \MB focuses on adaptively selecting replacement locations based on video motion characteristics.

Given a video sequence $x=\{x_t\}_{t=1}^{T}$, \MB first computes a motion intensity map based on adjacent-frame differences:
\begin{equation}
\mathcal{A}(h,w)=\frac{1}{3(T-1)}\sum_{t=1}^{T-1}\sum_{c=1}^{3}\big|x_{t+1}^{(c)}(h,w)-x_t^{(c)}(h,w)\big|,
\end{equation}
where $\mathcal{A}(h,w)$ reflects the average temporal variation at spatial location $(h,w)$.
This motion map provides a content-aware descriptor that characterizes the spatial distribution of temporal dynamics within the video.

The motion map $\mathcal{A}$ is then pooled into a patch-level motion distribution $\mathcal{P}$ using mask blocks, which size $b_0$.
Rather than adapting the spatial scale of replacement, \MB leverages motion statistics to guide the placement of the replacement budget.
Specifically, replacement locations are sampled according to an inverse-motion probability:
\begin{equation}
p_{ij}\propto 1-\mathcal{P}_{ij},
\end{equation}
to make sure that regions exhibiting lower temporal variation are more likely to be selected for replacement.
This strategy directs perturbations toward relatively stable regions, where replacement is less likely to disrupt critical dynamics or introduce temporal inconsistency.

Finally, by sharing the same spatial mask across all frames, \MB constructs a tube-consistent spatiotemporal mask. In particular, \uline{this tube-consistent masking ensures that replacement regions remain temporally aligned throughout the video, preserving structural continuity in the augmented sample}.
Through the above process, \uline{\MB introduces content-aware spatial and temporal constraints without additional supervision or trainable parameters}.

By integrating \MA and \MB, \ours provides a unified formulation of controlled mixing augmentation, in which statistical alignment and spatiotemporal adaptivity are jointly enforced.
Specifically, \MA regulates how much content is replaced and from which distribution the replacement originates, while \MB determines where and at what spatial scale the replacement budget is applied according to video motion characteristics.
Without introducing additional supervision or trainable parameters, \ours achieves stable representation expansion under controlled perturbations, leading to consistent performance gains across video behavior recognition tasks with varying temporal and spatial complexities.

\begin{table*}[t]
\renewcommand{\arraystretch}{1.0}
\centering
\setlength{\tabcolsep}{1.2mm}
\small

\scalebox{0.95}{
\begin{tabular}{ c|cc|cc|cc|c|cc }
\hline
\hline
\multirow{3}{*}{Method} & \multicolumn{2}{c|}{Coarse-grained} & \multicolumn{4}{c|}{Mid-grained}& \multicolumn{3}{c}{Fine-grained}  \\\cmidrule(lr){2-10}
 & \multicolumn{2}{c|}{UCF101}  & \multicolumn{2}{c|}{DFEW} & \multicolumn{2}{c|}{FERV39k} & \multicolumn{1}{c|}{MA52: Body} & \multicolumn{2}{c}{MA52: Action}   \\ 

\cmidrule(lr){2-10}
& Top-1 $\uparrow$  & Top-5 $\uparrow$   & WAR $\uparrow$  & UAR $\uparrow$   & WAR $\uparrow$ & UAR $\uparrow$  &  Top-1   &Top-1 & Top-5  \\ 
\hline
\hline
 \multicolumn{10}{l}{2D CNN based Methods }    \\
\cmidrule(lr){1-10}
ResNet &35.87 &62.57
&64.73 &55.22 
&45.24& 35.57 &60.04 &35.43 &73.52  \\
ResNet + \ours 
&\textbf{36.45${\uparrow}_{0.58}$} &\textbf{63.18${\uparrow}_{0.61}$}
&\textbf{66.48${\uparrow}_{1.75}$} &\textbf{57.84${\uparrow}_{2.62}$}
&\textbf{47.05${\uparrow}_{1.81}$} &\textbf{38.12 ${\uparrow}_{2.55}$} 
& \textbf{60.45${\uparrow}_{0.41}$}
&\textbf{35.89${\uparrow}_{0.46}$}
&\textbf{74.10${\uparrow}_{0.58}$}

 \\\cmidrule(lr){1-10}
ResNet\_LSTM 
&41.18&64.74
&67.08&54.63
&45.88& 37.14 
&60.59 &36.87 &73.88\\
ResNet\_LSTM + \ours &\textbf{42.80${\uparrow}_{1.62}$} &\textbf{65.05${\uparrow}_{0.31}$} 

&\textbf{67.64${\uparrow}_{0.56}$} &\textbf{55.62${\uparrow}_{0.99}$} 
&\textbf{47.18${\uparrow}_{1.30}$} &\textbf{ 38.28${\uparrow}_{1.14}$} 
&\textbf{61.24${\uparrow}_{0.65}$} &\textbf{37.63${\uparrow}_{0.76}$} &\textbf{74.20${\uparrow}_{0.32}$}
\\\cmidrule(lr){1-10}
 Average Improvement&$\uparrow$\textbf{1.10}&$\uparrow$\textbf{0.46} 
 &$\uparrow$\textbf{1.16}&$\uparrow$\textbf{1.81}
  &$\uparrow$\textbf{1.56}&$\uparrow$\textbf{1.85}
 &$\uparrow$\textbf{0.53} &$\uparrow$\textbf{0.61} &$\uparrow$\textbf{0.45}   \\\hline \hline
 \multicolumn{10}{l}{3D CNN based Methods}    \\
\cmidrule(lr){1-10}
R3D &62.69 &84.06 
&69.25 & 56.10 
&46.00&37.95  
& 72.68 &50.20 & 83.93\\
R3D + \ours &\textbf{63.91${\uparrow}_{1.22}$} &\textbf{84.95${\uparrow}_{0.89}$}
&\textbf{70.31 ${\uparrow}_{1.06}$} &\textbf{59.89 ${\uparrow}_{3.79}$} 
&\textbf{48.09${\uparrow}_{2.06}$} &\textbf{39.18${\uparrow}_{1.23}$}  
& \textbf{74.10${\uparrow}_{1.42}$} & \textbf{53.51${\uparrow}_{3.31}$} & \textbf{85.79${\uparrow}_{1.86}$}\\
\cmidrule(lr){1-10} 
 
X3D &64.26&85.67 
&67.21&57.76 
&46.16 &38.40 
&76.62 &51.52 & 84.64\\
X3D + \ours &\textbf{64.55${\uparrow}_{0.29}$}&\textbf{86.78${\uparrow}_{1.11}$} 
&\textbf{69.09${\uparrow}_{1.88}$}&\textbf{59.72${\uparrow}_{1.96}$}
&\textbf{48.67${\uparrow}_{2.51}$}&\textbf{39.72${\uparrow}_{1.32}$} 
& \textbf{77.87${\uparrow}_{1.25}$} &\textbf{54.30${\uparrow}_{2.78}$} &\textbf{86.54${\uparrow}_{1.90}$}
  \\\cmidrule(lr){1-10}
 Average Improvement &$\uparrow$\textbf{0.76}&$\uparrow$\textbf{1.00} 
 &$\uparrow$\textbf{1.47 }&$\uparrow$\textbf{2.88 }
 &$\uparrow$\textbf{2.29}&$\uparrow$\textbf{1.28}
 &$\uparrow$\textbf{1.34} &$\uparrow$\textbf{3.05} &$\uparrow$\textbf{1.88}  \\\hline \hline
  \multicolumn{10}{l}{Transformer based Methods}    \\
\cmidrule(lr){1-10} 
TimeSformer & 75.44&92.78 
&67.12 &57.13
& 47.11 &38.31 
&71.03 &44.70 & 83.51\\
TimeSformer + \ours &\textbf{76.71${\uparrow}_{1.27}$} &  \textbf{93.95${\uparrow}_{1.17}$}
&\textbf{67.94${\uparrow}_{0.82}$} &\textbf{59.68${\uparrow}_{2.55}$} 
&\textbf{48.21${\uparrow}_{1.10}$} &\textbf{40.01${\uparrow}_{1.70}$} 
&\textbf{72.18${\uparrow}_{1.15}$} &\textbf{44.96${\uparrow}_{0.26}$} & \textbf{84.04${\uparrow}_{0.53}$}\\\cmidrule(lr){1-10}
ViedoMAE &72.11 & 92.02
&69.35 &59.50 
&47.37 &38.62 
 &76.07 &55.67 &83.60 
\\
ViedoMAE + \ours &\textbf{73.91${\uparrow}_{1.80}$} & \textbf{92.44${\uparrow}_{0.40}$}
&\textbf{71.15${\uparrow}_{1.80}$} &\textbf{62.63${\uparrow}_{3.13}$}
&\textbf{47.99${\uparrow}_{0.62}$} &\textbf{39.33${\uparrow}_{0.71}$ }
 &\textbf{76.82${\uparrow}_{0.75}$} &\textbf{57.86${\uparrow}_{2.19}$} &\textbf{86.00${\uparrow}_{2.40}$} 
\\\cmidrule(lr){1-10}
 Average Improvement&$\uparrow$\textbf{1.54}&$\uparrow$\textbf{0.79}
 &$\uparrow$\textbf{1.31}&$\uparrow$\textbf{2.84}
 &$\uparrow$\textbf{0.86}&$\uparrow$\textbf{1.21}
 &$\uparrow$\textbf{0.95} &$\uparrow$\textbf{1.23} &$\uparrow$\textbf{1.47} 
 \\ \hline \hline
\end{tabular}}
\caption{Test comparisons (\%)  of different architectures on UCF101, DFEW, FERV39k, and MA52 datasets. (\textbf{Bold}: Best, \ul{Underline}: Second best.)}

\label{exp1}

\end{table*}

\begin{figure*}[t]   
\centering
\begin{minipage}{0.98\linewidth}
\centering
\begin{minipage}[t]{0.32\linewidth}
   \vspace{0pt}
  \centering
  \small
  \setlength{\tabcolsep}{1.8mm}
  \renewcommand{\arraystretch}{1.0}
  \scalebox{0.9}{
  \begin{tabular}{c|cc|cc}
  \hline\hline
  \multirow{2}{*}{Setting} & \multicolumn{2}{c|}{Method} & \multicolumn{2}{c}{DFEW}\\
  \cmidrule(lr){2-5}
  & \MA & \MB & WAR$\uparrow$ & UAR$\uparrow$\\
  \hline
  a & \ding{53} & \ding{53} & 67.21 & 57.76\\\cmidrule(lr){1-5}
  a & \ding{52} & \ding{53} & 68.71 & 58.26\\
  b & \ding{53} & \ding{52} & 68.79 & 57.40\\\cmidrule(lr){1-5}
  d & \ding{52} & \ding{52} & \textbf{69.09} & \textbf{59.72}\\
  \hline\hline
  \end{tabular}}
  \captionof{table}{Ablation (\%) study in \ours on DFEW dataset.}
  \label{exp1}
\end{minipage}\hfill
\begin{minipage}[t]{0.31\linewidth}
   \vspace{0pt}
  \centering
  \includegraphics[width=\linewidth]{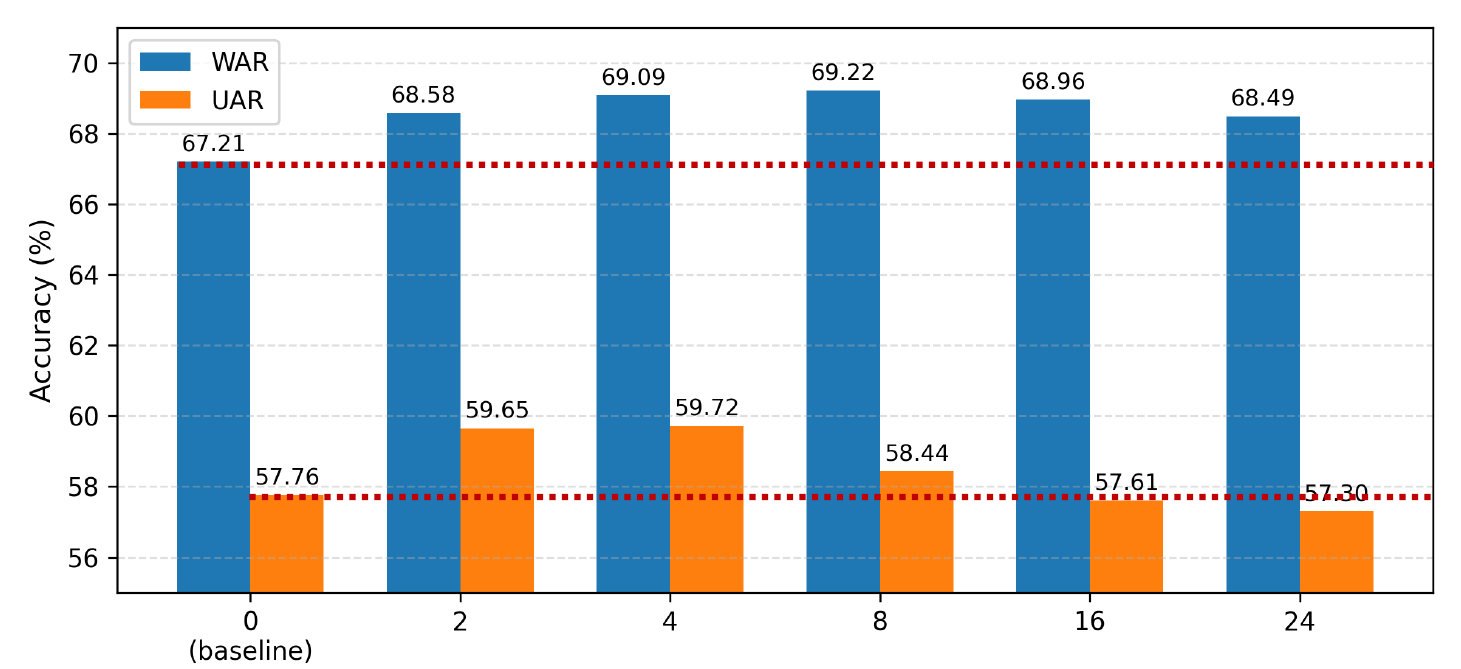}
  \captionof{figure}{Effect of different base block size on performance on DFEW.}
  \label{block}
\end{minipage}\hfill
\begin{minipage}[t]{0.31\linewidth}
   \vspace{0pt}
  \centering
  \includegraphics[width=\linewidth]{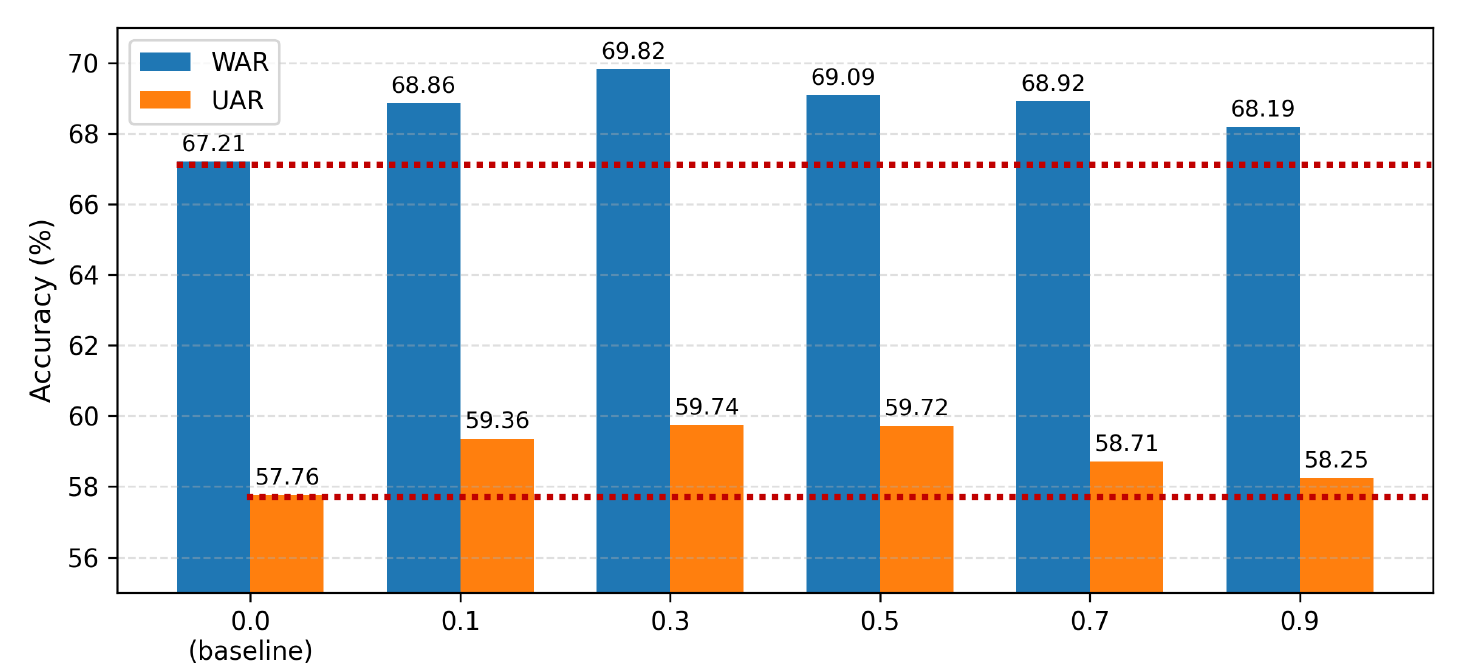}
  \captionof{figure}{Effect of different coverage ratio on performance on DFEW.}
  \label{mixing}
\end{minipage}
\end{minipage}
\end{figure*}

\begin{figure*}[t]
    \centering

    \begin{minipage}[t]{0.32\textwidth}
    \vspace{4pt}
        \centering
        \includegraphics[width=1.0\linewidth]{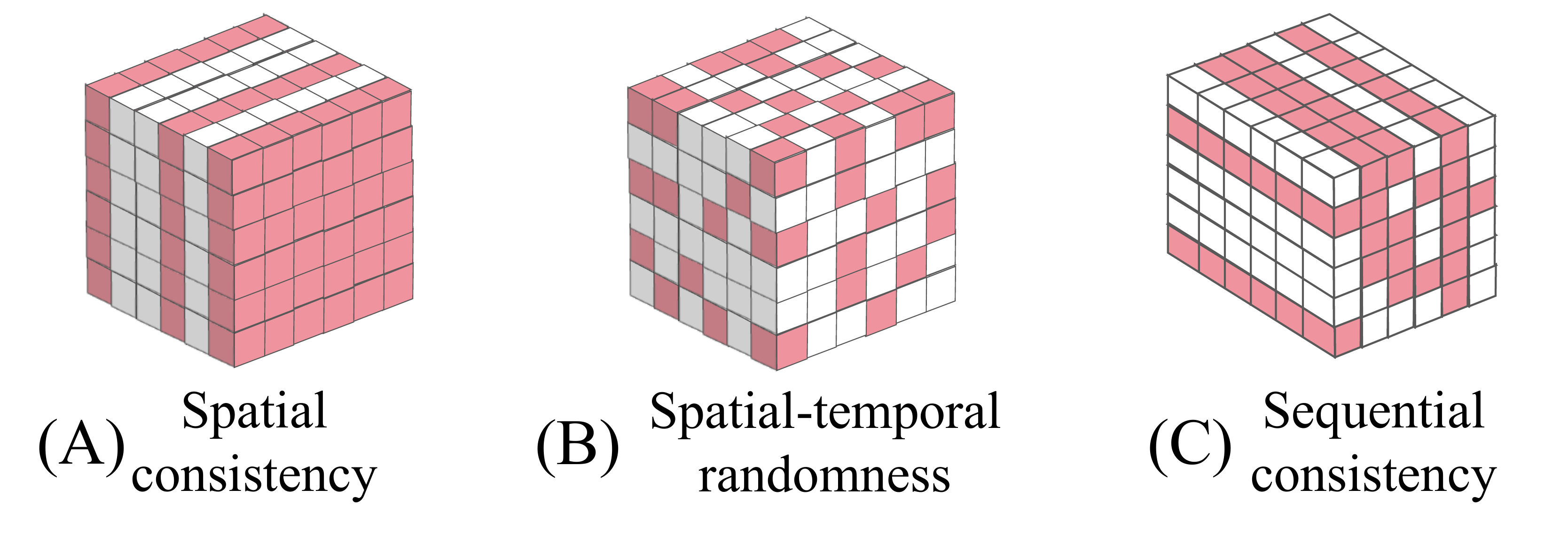}
        \captionof{figure}{Three types of spatiotemporal masking strategies.}
        \label{mixtype}
    \end{minipage}
    \hfill
    \begin{minipage}[t]{0.32\textwidth}
    \vspace{0pt}
        \centering
        \renewcommand{\arraystretch}{1.0}
        \setlength{\tabcolsep}{2mm}
        \small
        \scalebox{0.9}{
        \begin{tabular}{c|c|cc}
            \hline\hline
            Setting & Method & WAR$\uparrow$ & UAR$\uparrow$ \\
            \hline
            a & baseline & 67.21 & 57.76 \\\cmidrule(lr){1-4}
            b & A        & 67.29 & 58.03 \\
            c & B        & 68.24 & 57.12 \\\cmidrule(lr){1-4}
            d & C (\ours)& \textbf{69.09} & \textbf{59.72} \\
            \hline\hline
        \end{tabular}}
        \captionof{table}{Comparison (\%) of different spatiotemporal strategies on DFEW.}
        \label{typeexper}
    \end{minipage}
    \hfill
    \begin{minipage}[t]{0.32\textwidth}
    \vspace{0pt}
        \centering
        \renewcommand{\arraystretch}{1.0}
        \setlength{\tabcolsep}{2mm}
        \small
        \scalebox{0.9}{
        \begin{tabular}{c|c|cc}
            \hline\hline
            Setting & Method & WAR$\uparrow$ & UAR$\uparrow$ \\
            \hline
            a & baseline  & 67.21 & 57.76 \\\cmidrule(lr){1-4}
            b & mask only & 66.27 & 56.04 \\\cmidrule(lr){1-4}
            c & \ours     & \textbf{69.09} & \textbf{59.72}  \\
            \hline\hline
        \end{tabular}}
        \captionof{table}{Ablation (\%) study comparing mask-only and \ours on DFEW.}
        \label{maskonly}
    \end{minipage}

\end{figure*}

\section{Experiments}
\subsection{Experimental Setup}
\subsubsection{Datasets and Metrics}
We evaluate \ours on four different representative video benchmarks spanning coarse-, mid-, and fine-grained behavior recognition to assess its generalization across different motion scales and structural complexities.
UCF101 \cite{soomro2012ucf101} is used for coarse-grained body action recognition with large motion variations and is evaluated using Top-1 and Top-5 accuracy. 
DFEW \cite{dfew} and FERV39k \cite{ferv39k} focus on in-the-wild dynamic facial expression recognition with substantial appearance and intensity diversity, where Weighted and Unweighted Average Recall (WAR/UAR) are reported to account for class imbalance.
MA-52 \cite{guo2024benchmarking} is adopted for fine-grained micro-action recognition characterized by subtle motions and high intra-class similarity, and performance is measured by Top-1 and Top-5 accuracy.

\subsubsection{Implementation Details}
All experiments are implemented in PyTorch and conducted on a single NVIDIA RTX A6000 GPU. 
Training configurations are adjusted to accommodate different datasets and backbone architectures. 
For fair comparison, all methods sharing the same backbone are trained under exactly the same configuration.
\ours is applied only during training as a plug-and-play data augmentation strategy, without introducing additional trainable parameters or modifying backbone architectures. 

\subsection{Effectiveness on Different Datasets}

We evaluate \ours on multiple benchmark datasets spanning coarse-grained action recognition, in-the-wild facial expression recognition, and fine-grained micro-action recognition.
As a plug-and-play augmentation strategy, \ours is applied to diverse backbone architectures, enabling a comprehensive evaluation across different spatiotemporal characteristics and modeling paradigms.

\subsubsection{Overall Performance Analysis}
It is worth noting that across all datasets, \ours consistently improves recognition performance over the corresponding baselines.
These gains are observed across different backbone families, including 2D CNNs, 3D CNNs, and Transformer-based models, which clearly demonstrate that the effectiveness of \ours is backbone-agnostic and stems from improved data-level representation learning rather than architectural modifications.
By stabilizing feature learning under heterogeneous video distributions, \ours leads to more robust representations.

Notably, the improvements are particularly pronounced on balanced metrics such as WAR and UAR for facial expression datasets, where class imbalance and intra-class variability are prominent.
This suggests that \ours promotes more consistent intra-class representations and mitigates overfitting to dominant patterns, resulting in more stable and equitable performance across categories.

\subsubsection{Effectiveness Across Behavioral Granularities}
\ours consistently improves performance across behavior recognition tasks of different granularities, while its benefits manifest in task-specific ways.
On coarse-grained action recognition, \ours expands intra-class appearance and motion diversity under controlled replacement without disrupting global temporal semantics.
For facial expression recognition, which involves subtle dynamics and strong inter-subject variation, \ours yields notable gains in balanced metrics by combining statistically aligned intra-class mixing with motion-aware mask placement.
In fine-grained micro-action recognition, where discriminative cues are sparse and highly localized, \ours achieves particularly stable improvements by constraining perturbations to low-motion regions and enforcing temporal consistency, thereby preserving critical subtle motions.

\subsubsection{Consistency Across Backbone Architectures}
\ours also demonstrates consistent performance gains across different backbone architectures.
It complements convolution-based models by providing statistically aligned and structurally consistent training samples, and benefits Transformer-based models by mitigating representation noise caused by unconstrained perturbations.
Overall, results across multiple datasets and architectures indicate that \ours improves representation stability under heterogeneous spatiotemporal conditions, enabling more stable and discriminative representation learning.

\subsection{Ablation Studies}
We conducted a series of ablation studies on the DFEW dataset using X3D. DFEW has notable intra-class differences and class imbalance, and is designed to analyze the effects of the two core components in \ours.
\subsubsection{Ablation Study on Core Components of ReMA}
As reported in Tab.~\ref{exp1}, introducing either \MA or \MB alone improves performance over the baseline, while the gains remain limited when only a single mechanism is applied.

Removing \MA causes a more noticeable drop in UAR, indicating the importance of distribution-aware alignment for stabilizing class-level representations.
In contrast, removing \MB mainly degrades WAR, suggesting that motion-aware regulation is critical for preserving discriminative temporal cues.
When both components are jointly applied, \ours still achieves the best results on both WAR and UAR, which effectively confirms the complementary roles of \MA and \MB in enabling stable and effective augmentation under heterogeneous video conditions.

\subsubsection{Ablation on Spatiotemporal Mixing Consistency}

We further examine the role of spatiotemporal consistency by comparing three mixing strategies, \ie, spatially consistent mixing, spatiotemporally random mixing, and the temporally consistent (sequential) mixing adopted in \ours. The corresponding ablation results are shown in Fig.~\ref{mixtype} and Tab.~\ref{typeexper}.

Only temporally consistent mixing yields clear and stable improvements on both WAR and UAR.
Spatially consistent mixing provides marginal gains, while spatiotemporally random mixing leads to a noticeable degradation in UAR, indicating that frame-wise inconsistent perturbations disrupt intrinsic temporal statistics, especially under class imbalance.

By enforcing tube-level consistency across frames, temporally consistent mixing expands spatial diversity while preserving temporal structure, which effectively enables controlled replacement to operate on spatiotemporal redundancy without introducing artificial temporal noise.

\subsubsection{Effect of Base Block Size and Coverage Ratio}

We analyze the influence of two key hyperparameters in \ours, \ie, the base block size and the coverage ratio, which control the spatial granularity and overall strength of replacement, respectively, as shown in Fig.~\ref{block} and Fig.~\ref{mixing}.

Apparently, moderate base block sizes consistently yield the best performance.
Overly fine blocks introduce fragmented perturbations, whereas overly coarse blocks are more likely to disturb semantically meaningful regions and dominant motion structures.
Similarly, moderate coverage ratios achieve the most stable gains, \ie, small ratios provide insufficient variation, while large ratios weaken discriminative motion cues due to the excessive replacement.

These results confirm that the effectiveness of \ours relies on controlled augmentation, where spatial granularity and perturbation strength must be jointly regulated to balance diversity expansion and structural consistency.

 \begin{figure}[t]
\centering
\includegraphics[width=0.45\textwidth]{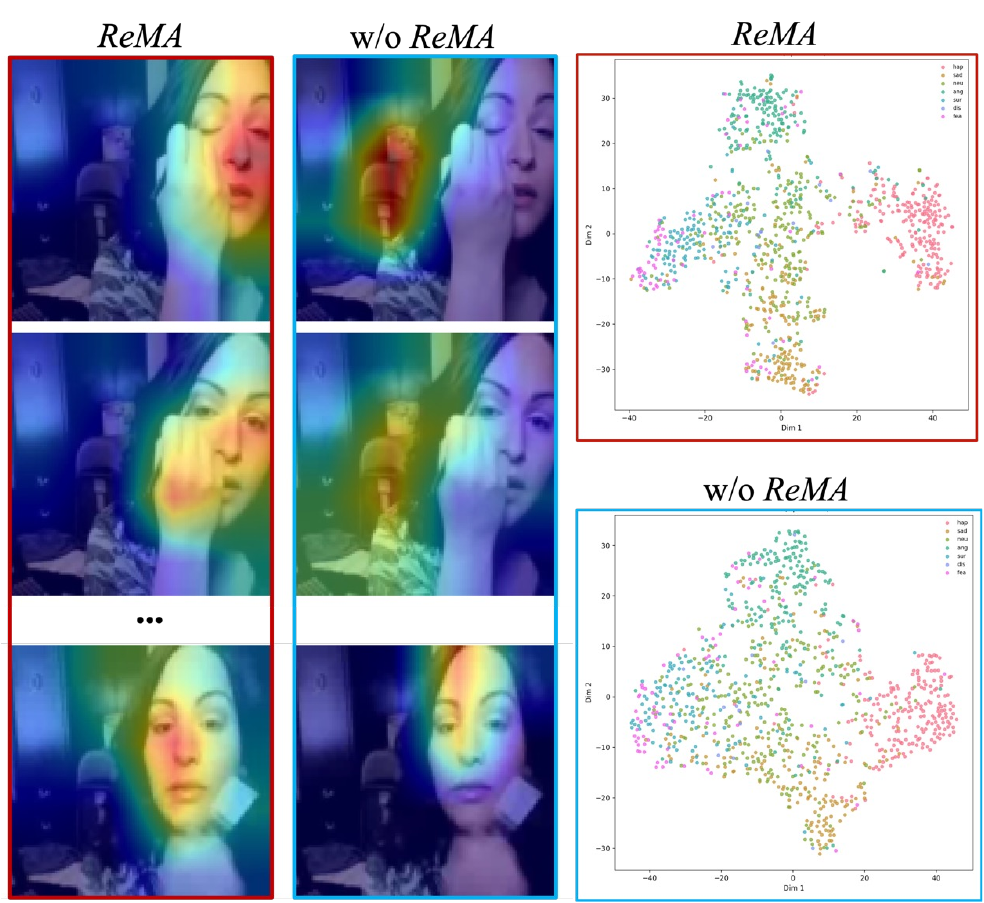}
\caption{Visualization of learned representations. }
\label{gridcam}
\end{figure}

\subsubsection{Ablation on Mask-only Augmentation}

We examine whether the performance gains of \ours arise from spatiotemporal masking itself by comparing it with a mask-only augmentation strategy, where masked samples are directly used for training without intra-class mixing~\cite{zhong2020random}.
The results are reported in Tab.~\ref{maskonly}.
Applying mask-only augmentation leads to a noticeable performance drop compared to the baseline.
This clearly indicates that masking alone mainly suppresses informative content and introduces irreversible information loss, without providing complementary intra-class variations to support representation learning, which is particularly detrimental under class imbalance.

In contrast, \ours significantly outperforms both the baseline and the mask-only setting.
This confirms that the effectiveness of \ours stems from controlled intra-class replacement regulated by spatiotemporal structure, rather than masking as a standalone operation.
In \ours, masking serves as a structural constraint to guide where mixing occurs instead of directly discarding information, enabling effective representation expansion through structured variability.

\subsection{Visualization}

Fig.~\ref{gridcam} presents qualitative visualizations of learned representations with and without \ours, including Grad-CAM activations on UCF101 (left) and t-SNE embeddings on DFEW (right).
Without \ours, activation maps are spatially scattered and temporally unstable, often responding to background regions or irrelevant textures, while feature embeddings from different categories exhibit substantial overlap.
These observations indicate that unconstrained augmentation can introduce representation drift and weaken class-conditional structure.

In contrast, models trained with \ours produce more compact and semantically coherent activation patterns that remain stable across frames, and their feature embeddings form tighter intra-class clusters with clearer inter-class separation.
This suggests that \ours effectively suppresses non-discriminative variations, enhances temporal consistency, and regularizes the representation space, which is consistent with the observed quantitative performance gains.
\section{Conclusion}
We present \ours, a representation-aware data augmentation method for video behavior recognition that addresses representation instability caused by spatiotemporal heterogeneity. By formulating augmentation as a controlled invariance process, \ours enables stable expansion of intra-class representations while preserving distributional structure.
\ours integrates two lightweight, plug-and-play components. \MA performs structured intra-class mixing to enhance discriminative diversity, while the \MB adaptively regulates spatiotemporal perturbations based on motion cues to maintain temporal coherence. Together, they form an effective data-level augmentation strategy without additional supervision or trainable parameters.
Extensive experiments across diverse datasets and backbone architectures demonstrate consistent improvements in robustness and generalization. Future work will explore extending \ours to broader spatiotemporal understanding tasks and more efficient deployment settings.





\bibliographystyle{IEEEtran}
\bibliography{icme2026references}

\end{document}